\def\year{2019}\relax
    \documentclass[letterpaper]{article}
    \usepackage{aaai19}
    \usepackage{times}
    \usepackage{helvet}
    \usepackage{courier}
    \usepackage{url}
    \usepackage{graphicx}
\usepackage{xcolor}
\usepackage{soul}
\usepackage[utf8]{inputenc}
\usepackage[small]{caption}
\usepackage{float}

\usepackage{amsmath}
\usepackage{comment}
\usepackage{float}
\usepackage{amsfonts}
\usepackage{booktabs}
\usepackage{siunitx}
\usepackage[linesnumbered,ruled]{algorithm2e}

\usepackage{amssymb}
\usepackage{times}

\newcommand\mafs{$\sf MAFS$}
\newcommand\mafsb{$\sf MAFSB$}
\newcommand\masiw{$\sf MA$-$\sf SIW$}
\newcommand\siw{$\sf SIW$}
\newcommand\iw{$\sf IW$}
\newcommand\bfws{$\sf BFWS$}
\newcommand\mabfws{$\sf MA$-$\sf BFWS$}
\newcommand\mabfwssecure{{\it secure}-$\sf MABFWS$}

\newcommand\maplan{$\sf MAPLAN$}
\newcommand\psm{$\sf PSM$}

\newcommand\blocksworld{$\rm Blocksworld$}
\newcommand\depot{$\rm Depot$}
\newcommand\driverlog{$\rm DriverLog$}
\newcommand\elevators{$\rm Elevators$} %'08
\newcommand\logistics{$\rm Logistics$} %'00
\newcommand\rovers{$\rm Rovers$}
\newcommand\satellites{$\rm Satellites$}
\newcommand\sokoban{$\rm Sokoban$}
\newcommand\taxi{$\rm Taxi$}
\newcommand\wireless{$\rm Wireless$}
\newcommand\woodworking{$\rm Woodworking$} %'08
\newcommand\zenotravel{$\rm Zenotravel$}
\newcommand\mabw{$\rm MA$-$\rm Blocksworld$}
\newcommand\mabwhard{$\rm MA$-$\rm Blocksworld$-$\rm Large$}
\newcommand\malog{$\rm MA$-$\rm Logistics$}
\newcommand\maloghard{$\rm MA$-$\rm Logistics$-$\rm Large$}
\newcommand\mabwshort{$\rm MA$-$\rm BW$}
\newcommand\mabwhardshort{$\rm MA$-$\rm BW$-$\rm L$}
\newcommand\malogshort{$\rm MA$-$\rm Log$}
\newcommand\maloghardshort{$\rm MA$-$\rm Log$-$\rm L$}

\usepackage[linesnumbered,ruled]{algorithm2e}

\newtheorem{definition}{Definition}
\newtheorem{theorem}{Theorem}

\newcommand{\tup}[1]{\langle #1\rangle}            % tuple
\SetKwProg{Fn}{Procedure}{}{}

\newif\ifLONGPAPER
\LONGPAPERfalse %save space

\title{Novelty Messages Filtering for Multi Agent Privacy-preserving Planning}
\author{Alfonso E. Gerevini \\ Dipartimento di Ingegneria dell'Informazione\\ Universit\`a degli Studi di Brescia, Italy\\
        \And Nir Lipovetzky \\ School of Computing and Information Systems \\ The University of Melbourne\\
        \AND Nico Peli \and Francesco Percassi \and Alessandro Saetti \and Ivan Serina\thanks{Corresponding author. Email: ivan.serina@unibs.it}  \\ Dipartimento di Ingegneria dell'Informazione \\ Universit\`a degli Studi di Brescia, Italy
}

\begin{document}

\maketitle
\begin{abstract}
In multi-agent planning, agents jointly compute a plan that achieves mutual goals, keeping certain information private to the individual agents. Agents' coordination is achieved through the transmission of messages. These messages can be a source of privacy leakage as they can permit a malicious agent to collect
information about other agents' actions and search states.
In this paper, we investigate the usage of novelty techniques in the context of (decentralised) multi-agent privacy-preserving planning, addressing the challenges related to the agents' privacy and performance. In particular, we show that the use of novelty based techniques can significantly reduce the number of messages transmitted among agents, better preserving their privacy and improving their performance. An experimental study analyses the effectiveness of our techniques and compares them with the state-of-the-art. 
Finally, we evaluate the robustness of our approach, considering different delays in the transmission of messages as they would occur in overloaded networks, due for example to massive attacks or critical situations. 
\end{abstract}

\section{Introduction}
Several frameworks for decentralised multi-agent (DMA) planning have been formalized and developed in the last few years, e.g., \cite{brafman2008one,nissim2014distributed,torreno2014fmap}.  An important issue of these approaches is related to how they handle agents' privacy; in fact, for DMA planning agents have private knowledge that they do not want to share with others during the planning process and  plan execution. This issue prevents the straightforward usage of most of the modern techniques developed for centralized (classical) planning, which are based on heuristic functions computed by using the knowledge of all the involved agents.

In DMA planning, computing search heuristics using the knowledge of all the involved agents may require many exchanges of knowledge among agents, and this may compromise the agents' privacy. For preserving the privacy of the involved agents, the distance from a search state to the goal states can be estimated by using either the knowledge of a {\em single} agent alone, or the public projections of actions of the involved agents. However, this estimate is much more inaccurate than using the knowledge of all the agents.

Given that for classical planning best-first width search performs very well even when the estimate of the goal distance is inaccurate \cite{LipovetzkyG12}, such a procedure represents a good candidate to effectively solve MA-planning problems without compromising the agents' privacy.
Recently, for MA planning \citeauthor{GereviniMAicaps19}   (\citeyear{GereviniMAicaps19}) proposed an effective search procedure called \mabfws\ that uses width-based exploration in the form of novelty-based preferences to provide a complement to goal-directed heuristic search. 
In order to preserve the privacy of the involved agents, the private knowledge shared among agents is encrypted. An agent $\alpha_i$ shares with the other agents a description of every reached search state in which all the private facts of $\alpha_i$ that are true in a state are substituted with a string obtained by encrypting all the private fact names of $\alpha_i$ together. Notably, this encryption has an impact on the measure of novelty, and hence also affects the definition of the heuristic guiding the search \cite{GereviniMAicaps19}. 

In this paper we investigate the use of novelty to filter the messages sent by each agent, we propose different methods to exploit such filtering within forward search MA planning, and we discuss its properties in terms of privacy.
The following sections present the background on width-based search and privacy in MA planning, discuss related work, propose our filtering techniques based on the notion of novelty, show the results of an experimental study to evaluate the effectiveness of the proposed novelty message heuristics, and finally give the conclusions and mention future work.

\section{Background}\label{sec:background}

In this section, first we introduce the MA-STRIPS planning problem, then we describe some prominent width-based search procedures developed for classical planning.

\subsection{The MA-STRIPS planning problem}

Our work relies on MA-STRIPS, a ``minimalistic'' extension of the STRIPS language for MA planning \cite{brafman2008one}, that is the basis of the most popular definition of MA-planning problem (see, e.g., \cite{nissim2014distributed,brafman15privacy,maliah2014privacy,ppLAMA,nissim2012multi}). %nissim2012tunneling

\begin{definition}
A  {\sc MA-STRIPS} {planning problem} $\Pi$ for a set of agents $\Sigma = \{\alpha_i\}_{i = 1}^n$ is a 4-tuple $\langle \{A_i\}_{i=1}^n,~ P,~I,~G  \rangle$ where:

%\vspace{-2mm}
\begin{itemize}
\item $A_i$ is the set of actions agent $\alpha_i$ is capable of executing, and s.t.\
for every pair of agents $\alpha_i$ and $\alpha_j$ $A_i \cap A_j = \emptyset$;   
\item $P$ is a finite set of propositions;
\item $I \subseteq P$ is the initial state;
\item $G \subseteq P$ is the set of goals.

\end{itemize}
\end{definition}

\noindent 
Each action $a$ consists in a name, a set of preconditions, $Prec(a)$, representing facts required to be true for the execution of the action, a set of additive effects, $Add(a)$, representing facts that the action makes true, a set of deleting effects, $Del(a)$, representing facts that the action makes false, and a real number, $Cost(a)$, representing the cost of the action.
A fact is {\it private} for an agent if other agents can neither
achieve, destroy nor require the fact \cite{brafman2008one}. A fact is {\it public}
otherwise.   An action is {\em private} if all its preconditions and
effects are private; the action is {\em public}, otherwise. A state obtained by executing a public action is said to be public; otherwise, it is private.
In the rest of the paper, we denote by $public(s)$ the public part of a state $s$.

To maintain agents' privacy, the private knowledge shared among agents can be encrypted. An agent can share with the other agents a description of its search state in which each private fact that is true in a state is substituted with a string obtained by encrypting the fact name \cite{Bonisoli2018}. This encryption of states does not reveal the names of the private facts of each agent $\alpha_i$ to other agents, but an agent can realize the existence of a private fact of agent $\alpha_i$ and monitor its truth value during search. This allows the other agents to infer the existence of private actions of $\alpha_i$, as well as to infer their causal effects. Another way of sharing states containing private knowledge during the search is to substitute, for each agent $\alpha_i$, all the private facts of $\alpha_i$ that are true in a state with a string obtained by encrypting all private fact names of $\alpha_i$ together \cite{nissim2014distributed}. Such a string denotes a dummy private fact of $\alpha_i$, which is treated by other agents as a regular fact. The work presented in this paper uses this latter method for the encryption of states. With this method,  other agents can only infer the existence of a group of private facts of $\alpha_i$, since the encrypted string contained in the states exchanged by $\alpha_i$ substitutes a group of an arbitrary number of private facts of $\alpha_i$. 

\ifLONGPAPER
A popular algorithm for solving MA-STRIPS planning problem is \mafs\ \cite{nissim2014distributed}, the distributed variant of forward best-first search. Essentially, in \mafs\ each agent considers a separate search space: each agent maintains an its own open list of states that are candidates for expansion and an its own closed list of already expanded states. Each agent expands the state among those in its open list which is estimated to be most promising for reaching the problem goals. When an agent expands a state, it uses only its own actions. If the action used for the expansion is public, the agent sends a message containing the expanded public state to other agents. When an agent receives a state via a message, it checks whether this state appears in its open or closed lists. If it is not contained in these lists, the agent inserts the state into its open list. 
\fi

\subsection{Privacy} 
Privacy in MA-STRIPS planning is concerned with guaranteeing that the private information of an agent $\alpha_i$  remains known only by agent $\alpha_i$, namely: its private variables and  values, as well as the existence, structure and cost of its private actions.  \citeauthor{brafman15privacy} (\citeyear{brafman15privacy}) defines an algorithm as \emph{weakly privacy} preserving if no value of private variables is shared with other agents, and the information shared among agents consists of  \emph{public projections} of public actions, i.e. private preconditions and effects are dropped from public actions. Algorithms for MA planning typically achieve weak privacy by encrypting the private variables of a state before sending it.

Weakly private algorithms allow agents to track changes in states sent by one agent and deduce the existence of private variables. In contrast, an algorithm is \emph{strongly private} if no agent $\alpha_i$ can deduce about the existence of i) a value or variable private to agent $\alpha_j$, $i\neq j$, and ii) the model of private actions of $\alpha_j$,  beyond 1) what its own actions $A_i$ reveals, 2) the public projection of actions $A_j$, and 3) the public projection of the actions in the solution plan.

One way to hinder the deduction ability of an agent is randomizing the order of exchanged messages, which can be achieved by delaying messages with random times from a given probabilistic distribution. This has not been studied formally, but we experimentally evaluate the impact of delays in MA planning algorithms.

\citeauthor{brafman15privacy} (\citeyear{brafman15privacy}) also proposes \emph{secure-MAFS}, a complete and sound forward search algorithm that achieves \emph{strong privacy} when the heuristic used is independent from the private part of the problem and all actions have unary cost. The key insight relies on making sure that an agent sends a state $s$ to other agents iff the public projection of state $s$ has never been sent before. This ensures that agents never receive two states with the same public projection from the same source agent, which is sufficient to guarantee that agents cannot distinguish between the execution of \emph{secure-MAFS} among $\Pi, \Pi' \in C$, where $C$ is an equivalence class containing all problems that share the same public solution space as the original problem $\Pi$ being solved \cite{brafman15privacy,tovzivcka2017limits}. Other notions of privacy have been proposed; e.g., cardinality privacy prevents an agent from inferring the number of private objects of the same type managed by other agents \cite{maliah2016stronger}.

%\vspace*{3mm}
\subsection{Width-based Search}
Pure width-based search algorithms are exploration algorithms that do
not pay attention to the problem goals at all. The simplest of such algorithm is $IW(1)$,
which is a plain breadth-first search where newly generated states
that do not make an atom $X = x$ true for the first time in the search
are pruned. The algorithm $IW(2)$ is similar except that a state $s$
is pruned when there are no atoms $X = x$ and $Y = y$ such that the
pair of atoms $X = x$, $Y = y$ is true in $s$ and false in all the
states generated before $s$. 

$IW(k)$ is a standard breadth-first search except
that newly generated states $s$ are pruned when their ``novelty'' is
greater than $k$, where the novelty of $s$ is $i$ iff there is a tuple
$t$ of $i$ atoms such that $s$ is the first state in the search that
makes all the atoms in $t$ true, with no tuple of smaller size having
this property \cite{LipovetzkyG12}.  While simple, it has been shown
that $IW(k)$ manages to solve arbitrary instances of many of the
standard 
%single agent 
benchmark domains in low polynomial time
provided that the goal is a single atom. Such domains can be shown to
have a small and bounded {\em width} $w$ that does not depend on the
instance size, which implies that they can be solved (optimally) by
running $IW(w)$. Moreover, $IW(k)$ runs in time and space that are
exponential in $k$ and not in the number of problem variables.  

The procedure $IW$, that calls $IW(1)$ and $IW(2)$,
sequentially, has been used to solve instances featuring multiple
(conjunctive) atomic goals, in the context of Serialized IW
(\siw) \cite{LipovetzkyG12}, an algorithm that calls $IW$ for
achieving one atomic goal at a time.  In other words, the $j$-th
subcall of \siw\ stops when IW generates a state $s_j$ that consistently
achieves $j$ goals of $G$.  The  state $s_j$ {\em consistently
achieves} $G_j \subseteq G$ if $s_j$ achieves $G_j$, and $G_j$ does
not need to be undone in order to achieve $G$. This last condition is
checked by testing whether $h_{max}(s_j) = \infty $ is true once the
actions that delete atoms from $G_j$ are excluded. 
While \siw\ is an incomplete blind search 
procedure (if dead-ends exist),
it turns out to perform better than a greedy best-first
search guided by standard delete relaxation heuristics  
\cite{LipovetzkyG12}.

Width-based exploration in the form of simple novelty-based preferences instead of pruning can provide an effective complement to goal-directed heuristic search without losing completeness. Indeed, it has been recently shown that for classical planning the combination of width-based search and goal-directed heuristic search, called best-first width search (\bfws), yields a search scheme that is better than both, and outperforms the state-of-the-art planners \cite{DBLP:conf/aaai/LipovetzkyG17}. 

%%\ifLONGPAPER
\bfws$(f)$ with $f = \langle h, h_1 , ... , h_n\rangle$ is a standard
best-first search that uses the function $h$ to rank the nodes
in the {\em open list}, breaking ties lexicographically with $n$ functions
$h_1, \dots, h_n$. The primary evaluation function $h$ is given by
the novelty measure of the node. To integrate novelty with goal directed heuristics, the notion of novelty used by \bfws\ is different from that used for the breadth-first search IW. For \bfws, given the functions $h_1, \dots, h_n$,  the novelty $w(s')$ of a
newly generated state $s'$  is $i$
iff there is a tuple (set) of $i$ atoms $X_i = x_i$ and no tuple of
smaller size, that is true in $s$ but false in all previously 
generated states $s'$ with the same function values $h_1 (s' ) = h_1 (s),
\dots,$ and $h_n (s' ) = h_n (s)$. 
%%\fi
For example, a new state $s$ has novelty $1$ if there is an atom $X = x$ that is true in $s$ and false in all the states $s'$ generated before $s$ where $h_i (s' ) = h_i (s)$ for all $i$. In the rest of the paper, the novelty measures $w$ is sometimes denoted as $w_{(h_1,\dots,h_n)}$  in order to make explicit the functions $h_1,\dots, h_n$ used in the definition  and computation of $w$.

\section{Related Work}

A MA-planning algorithm similar to ours is \mafs\ \cite{nissim2014distributed}. %\cite{nissim2012multi,nissim2013cost,nissim2014distributed}. 
%As discussed before, 
\mafs\ is a distributed best first search that for each agent considers a separate search space.
\ifLONGPAPER
\citeauthor{MAFSB} (\citeyear{MAFSB}) propose \mafsb, an enhancement of \mafs\ that uses a form of backward messages to reduce the number of search states shared by the agents. With our approach, the number of exchanged states can be limited by pruning the states with a novelty greater than a given bound.  
\fi
The existing work investigating the use of a distributed A* for partial-order MA-planning shares our motivations  \cite{torreno2014fmap}.
Differently from this approach, our MA-planning procedure searches in the space of world states, rather than in a space of partial plans, and it exchanges states among agents rather than partial plans.

\ifLONGPAPER
Different heuristics are proposed by \cite{maliah2014privacy,stolba2015admissible,torreno2014fmap}. They are landmarks-based heuristics, and a heuristic that is based on the domain transition graph, augmented by a special node in the graph that represents when the agent does not know the whole domain of a variable. Moreover, \citeauthor{torreno2015global} (\citeyear{torreno2015global}) show that a hybrid approach, i.e. that alternately uses two different heuristics to evaluate the generated states, can be more effective in terms of performance.  When computing the heuristic function, in the worst case every agent is involved, and this can make the computation slow because of a potentially very heavy communication overhead. Current research is focusing on avoiding too much (costly) communication among the agents. The most promising heuristics in this scope are ``potential heuristics'' \cite{stolbapotential,stolbacost}. 
\fi

Using width-based search for MA planning is not a novel idea. 
\iw\ search was used for solving a classical planning problem obtained from the compilation of a multi-agent planning problem \cite{Muise2015}. This work is for centralized MA planning, while our work investigates the distributed MA planning problem. \citeauthor{bazzotti2018iterative} (\citeyear{bazzotti2018iterative}) study the usage of Serialized-IW (abbreviated by \masiw) in the setting of distributed MA planning. The MA problem solved by \masiw\ is split into a sequence of episodes, where each episode $j$ is a subproblem solved by IW, returning a path to a state where one more problem goal has been achieved with respect to the last episode $j-1$.
Finally, \citeauthor{GereviniMAicaps19} (\citeyear{GereviniMAicaps19}) study the properties of \mabfws\ in terms of the width of MA-STRIPS problems, and propose several evaluation functions combining novelty and new heuristic functions where state-of-the-art performance is achieved without sharing the public projection of actions across agents. Differently from these approches using width-based search, in this work we study the usage of novelty to filter the messages sent by each agent, instead of to prune or evaluate search states.

\ifLONGPAPER
For this, during the search each agent sends two type of messages to other agents, messages containing a (public) state, and restart messages, i.e., message containing a (public) state from which starting a new search episode. The episode of an agent can terminate if it achieves a state with a new goal, or if it receives a restart message from another agent.      
\fi

\section{On the use of Novelty for message transmission}
The aim of our work is to reduce the number of search states
exchanged among agents during the search phase. Each agent retains,
therefore does not send to  other agents, the search states whose
novelty value called \textbf{outgoing novelty} exceeds a given threshold. The outgoing novelty is computed considering the public part of the search states previously transmitted to the other agents.
\begin{definition}
\label{def:outnovelty}

The \textbf{outgoing novelty} of a state $s$ 
given $m$ functions $h_1 , \dots , h_m$ is $k$ (denoted as $w^{out}_{h_1,...,h_m}(s)=k$) iff there is a tuple
(conjunction) of $k$ atoms and no smaller tuple, that is true in the {\em public part} of $s$ and false in the {\em public part} of all states $s'$ previously transmitted  with the
same function values, i.e., with $h_i (public(s')) = h_i(public(s))$ for $1 \le i \le m$. If no such tuple exists, then $w^{out}_{h_1,...,h_m}(s)$ is set to the maximum value $|P^{pub}|+1$, where $|P^{pub}|$ is the number of public propositions in the problem. 
\end{definition}

Based on the outgoing novelty value, a state $s$ can be withheld or transmitted to other agents.

\begin{definition}
The \textbf{withheld states} are the states that have not been sent to the other agents because of the fact that their outgoing novelty exceeds a given threshold.
\end{definition}

\noindent
In order to preserve the completeness of the algorithm, these states can be
transmitted to the other agents under specific conditions.
To describe these
conditions, it is necessary to introduce the definition
of \textbf{partially empty search}. 
\begin{definition}\label{def:partiallyEmpty}
The search of
an agent $\alpha$ is \textbf{partially empty} when the following conditions hold:
\begin{enumerate}
    \item the {\em open list} of $\alpha$ is empty;
    \item $\alpha$ has no other entry message to process;
    \item $\alpha$ has at least one withheld state.
\end{enumerate}
If condition 1 and 2 hold but condition 3 does not hold, then the search of $\alpha$ is {\bf empty}.
\end{definition}

An agent whose search phase is empty or partially empty is called ``\emph{waiting}''.
When an agent $\alpha$ is {\em waiting}, it can send a part of, or
even all, its withheld states (if any) to the other agents in order
to ``reinvigorate'' the search process of the other agents, or it can ask
the other agents to send their withheld states to it in order to restore its search process. In contrast, $\alpha$ could wait that
at least a given number of agents, indicated with {\em
$num\_waiting$}, are in the same condition before sending its withheld states
or asking the withheld states of the other agents. With this purpose, the agents communicate
if their search is empty or partially empty
to the others; when the number of waiting agents exceeds value $num\_waiting$, the agents transmit their withheld
states.\footnote{The transmission of the message ``empty search space''
 among the agents is necessary in order to allow the agents
to terminate their search before the timeout. }
In particular, we distinguish among three situations: 
\begin{itemize}
    \item $num\_waiting=1$, agents transmit/request their withheld states when at least one agent is waiting;
    
    \item $num\_waiting=half$, agents transmit/request their withheld states when at least half the agents in the problem is waiting;
    \item $num\_waiting=all$, agents transmit/request their withheld states when all the agents in the problem is waiting.
    
\end{itemize}

As previously explained, the waiting agents can transmit/request
withheld states to the other agents. Both these situations can heavily
influence the search process. We considered three
configurations, indicated with $who\_send$, in order to
define which agents send the withheld states. In particular when

\begin{itemize}
    \item $who\_send=waiting$, the \emph{waiting} agents send their withheld states; 
    \item $who\_send=not\; waiting$, the \emph{not waiting} agents send their withheld states; 
    \item $who\_send=all$, all the agents send their withheld states.
\end{itemize}
The best configuration seems to be the second one
($who\_send=not\; waiting$), because since the waiting agents are idle and
need states from the other agents to restore their search.
On the other hand, with $who\_send=waiting$, the
waiting agents can take advantage of the period of 
inactivity to send their withheld states.

When the previous conditions are verified the agents can send all their withheld states or only a part of them.
Specifically, we considered four different configurations, indicated with $num\_withheld\_states$, that specifies which states, among the withheld ones, have to be sent when necessary:
\begin{itemize}
    \item $num\_withheld\_states=none$, no state is sent (completeness is lost);
    \item $num\_withheld\_states=1$, one state at the time is sent (the one with lowest heuristic value);
    \item $num\_withheld\_states=group$, all withheld states with the lowest heuristic value are sent;
    \item $num\_withheld\_states=all$, all withheld states are sent.
\end{itemize}

%\vspace*{3mm}
\subsection{Privacy, Soundness and Completeness}

We study the theoretical properties of novelty-based message filtering for sound and complete MA forward search planners. Without loss of generality, we focus on \mabfws, which is weakly privacy preserving: it does not require sharing the public projection of public actions, and sends messages containing only descriptions of states obtained by encrypting private facts. 

\begin{theorem}
\mabfws\ with novelty messages filtering is sound and complete, iff  $num\_withheld\_states\neq none$. 
\end{theorem}

\noindent
{\bf Proof sketch.}
If $num\_withheld\_states\neq none$, then eventually all public states are going to be sent. $num\_waiting$ and $who\_send$ only changes the order of the exchanged messages, but does not preclude that messages will be sent later in the search.
$\Box$\\

Following previous definitions of strong privacy \cite{brafman15privacy,tovzivcka2017limits}, given a problem $\Pi$, $C$ is the set of problems $\Pi'$ where their public projection is equivalent, i.e., $\Pi^{pub} = \Pi'^{pub}$ and the set of public projections of reachable public states for $\Pi$ is the same as for $\Pi'$. Note that $\Pi$ and $\Pi'$ differ only in their private parts. \mabfws\ is strong privacy preserving with respect to the class $C$ of MA problems, iff all agents cannot distinguish the execution of \mabfws\ with input $\Pi$ and $\Pi' \in C$.  An agent cannot distinguish iff the number of messages  sent by an agent is the same for both problems, and the number of these messages does not depend on the private part of the problem \cite{tovzivcka2017limits}.

\begin{theorem}
\label{th:strong}
\mabfws\ is incomplete but strong privacy preserving for the problems in class $C$, for any threshold $k \in \{0,..,|P^{pub}|\}$ when used with:
\begin{enumerate}
    \item $num\_withheld\_states = none$, namely pruning messages with outgoing novelty greater than the threshold;
    \item no function $h_i$ is used in Definition~\ref{def:outnovelty};
    \item the heuristic functions used to guide the search are agnostic of the cost and private propositions of the agents.
\end{enumerate}
\end{theorem}

\noindent
{\bf Proof.}
We assume MA-BFWS encrypts the private part of $s$ and no information is leaked from the communication. By Definition~\ref{def:outnovelty}, for any value of outgoing novelty, a state $s$ is sent iff no other state $s'$ with the same public projection has been sent before. The number of sent messages depends only on the public part of the problem, as the private part is not taken into account in the computation of outgoing novelty. As both $\Pi$ and $\Pi'$ share the same public problem and reachable public state space, the number of messages sent will be equivalent. Therefore, it is strong privacy preserving. 
$\Box$\\

The state expansion order of \mabfws\ is independent from the private part of the problem if the search is guided, for example, by $f = \langle w_{(\#g)}, d\rangle$, where $w_{(\#g)}$ is the novelty over the public projection of the states using  a goal counting heuristic $\#g$, breaking ties with the depth $d$ of the public actions leading to the current state.
The version, with $num\_withheld\_states = none$, is incomplete if a state whose public projection has been sent before needs to be sent again in order to find a solution. \mabfws\ can be made complete by using a strategy similar to that proposed for secure-MAFS \cite{brafman15privacy}. 

When strong privacy cannot be preserved, reducing the number of exchanged messages is a good strategy to decrease the deductive capability of other agents. This can be accomplished by filtering messages according to the notion of novelty. Indeed, this filtering makes sure that messages with different state variable values are sent first, before sending states with repeating values that can lead to information leakage.

%\vspace*{3mm}
\section{Experiments}

In this section, we present an experimental study aimed at testing the effectiveness of the filtering techniques described so far. First, we describe the experimental setting; then we evaluate the effectiveness of our heuristics and compare the performance of our approach with the state-of-the-art; finally, we experiment different delays in the transmission of messages.

%\subsection{Experimental settings}

\begin{table}[t]
\centering
\small
\begin{tabular}{|l|c|c|c|}
\hline
Performance & $nw=all$
& $nw=1$ 
& $nw=half$ \\ 
	\hline 
\bf{Coverage} (320)&259&279&{\bf 280} \\ 
Avg time&{\bf 5.01}&8.57&5.13 \\ 
Avg quality&185.89&185.79&{\bf 184.75} \\ 
k-messages&{\bf 75.32}&543.43&76.69 \\ 
k-states&144.26&308.11&{\bf 144.22} \\ 
IPC Quality&222.81&239.46&{\bf 240.31} \\ 
IPC Time&241.68&252.42&{\bf 258.06} \\ 
Stdev Time&{\bf 0.53}&2.8&1.01 \\ 
Stdev Quality&{\bf 15.18}&16.72&15.19 \\ 
	\hline 
\end{tabular}
\caption{Performance of \mabfws\ with $num\_waiting$ (abbreviated with $nw$) set to 1, $half$, and $all$ in terms of number of solved problems, average CPU time (in seconds), average plan cost, number of sent messages (in thousands), number of expanded states (in thousands), IPC time, IPC quality, standard deviation of the CPU times, and standard deviations of the plan costs. The best performances are indicated in bold.}\label{tab:resultsfinaliS}
\end{table}

Our code is written in C++, and exploits the Nanomsg open-source library to share messages \cite{nanomsg}.\footnote{Our code and experimental data will be made available.}
 Each agent uses three threads, two of which send and receive messages, while the other one conducts the search, so that the search is asynchronous w.r.t.\ the communication routines. The behaviour of \mabfws\ depends on the order with which the messages are received by an agent. Each time a run of \mabfws\ is repeated, the agents' threads can be scheduled by the operative system differently, so that the behaviour of \mabfws\ can also be different. Thereby, for each problem of our benchmark, we run \mabfws\ five times and consider the performance of the algorithm as the median over the five runs. When \mabfws\ exceeds the CPU-time limit for more than two of the five runs, we consider the problem unsolved.

In our experiments, \mabfws\ search uses  \mabfws\ uses heuristic $f_6 =  \tup{w_{(G_\bot, \#r)},G_\bot,\#r}$, where novel search states are expanded first, breaking ties with counters based on the goals $G_\bot$ and a relaxed plan $\#r$ computed once from the initial state. $G_\bot$ counts the number of unachieved goals, and $\#r$ counts how many facts from a single relaxed have been achieved on the way to a state $s$. For more details the interested reader can  see the work by  \citeauthor{GereviniMAicaps19} (\citeyear{GereviniMAicaps19}).  The benchmark used in our experiments includes twelve domains proposed by  \citeauthor{vstolba2015competition} (\citeyear{vstolba2015competition}) for the distributed track of the first international competition on distributed and multi-agent planning (CoDMAP), and four domains \mabw\ (shortly, \mabwshort), \mabwhard\ (\mabwhardshort), \malog\ (\malogshort), \maloghard\ (\maloghardshort), which were proposed by \citeauthor{MAFSB17} (\citeyear{MAFSB17}). In the following, these latter four domains are abbreviated to MBS. The difference w.r.t.\ the CoDMAP domains \blocksworld\ and \logistics\ is that for the domains of MBS many private actions need to be executed between two consecutive public actions, some goals can be private, and agents must choose among several paths to achieve goals. %{ All domains have uniform action costs}. 

All tests were run on an Ubuntu server with 24 cores  Intel Xeon E5-2620 with 2 GHz and 128 Gbytes of RAM. Given a MA-planning problem, for each agent of the problem we limited the usage of resources to 3 CPU cores and 8 GB of RAM. Moreover, unless otherwise specified, the time limit was 5 minutes, after which the termination of all threads was forced.

\subsection{Novelty filtering}

\begin{table}[t]
\centering
\small
\begin{tabular}{|l|c|c|c|}
\hline
Performance & $ws=wait$
& $ws=not\;wait$  
& $ws=all$   \\  
	\hline 
\bf{Coverage} (320)&279&278&{\bf 280} \\ 
Avg time&5.59&6.04&{\bf 5.5} \\ 
Avg quality&180.0&{\bf 178.61}&179.64 \\ 
k-messages&{\bf 88.7}&109.72&92.79 \\ 
k-states&227.6&228.48&{\bf 178.97} \\ 
IPC Quality&238.03&238.58&{\bf 240.39} \\ 
IPC Time&{\bf 259.76}&257.82&259.21 \\ 
Stdev Time&{\bf 1.14}&1.73&1.27 \\ 
Stdev Quality&16.88&14.86&{\bf 14.72} \\ 
	\hline  
\end{tabular}
\caption{Performance of \mabfws\ with $num\_waiting=half$, and $who\_send$ (abbreviated with $ws$) set to $wait$, $not\,wait$, and $all$ in terms of number of solved problems, average CPU time (in seconds), average plan cost, number of sent messages (in thousands), number of expanded states (in thousands), IPC time, IPC quality, standard deviation of the CPU times, and standard deviations of the plan costs. The best performances are indicated in bold.}\label{tab:resultsfinaliX}
\end{table}

The same measure of novelty $w_{(G_\bot, \#r)}$ used for guiding the search is used for filtering messages. We denote by $w^{out}$ the maximum value of outgoing novelty a state $s$ can have to not be withheld.
We experimentally study different conditions under which an
agent can decide to transmit (a part of) its withheld
states. In our experiments, unless differently specified, \mabfws\ is used with $num\_waiting=half$,  $who\_send=all$, $num\_withheld\_states=group$, and $w^{out}=1$. The latter condition means  that states with outgoing novelty greater than one have been withheld. 

In the rest of the paper, for each experiment the average values are computed over the problems solved by all the compared approaches. The lower the average plan quality, the better the performance. Time and quality scores are measured by the score functions used for the seventh international planning competition. Higher values indicate better performance.\footnote{ For details on the scores of the seventh planning competition, see \url{http://www.plg.inf.uc3m.es/ipc2011-learning}.} 

The first experiment we conducted concerns the decision about when an agent sends its withheld states. 
The results in Table \ref{tab:resultsfinaliS} show that $num\_waiting=all$ is the
configuration that sends the fewest states, but it is also the one with the far
lowest coverage. This is probably due to the fact that requiring
that all agents are in waiting state is a condition
that reduces the transmission of withheld states too much. 
%Instead with $num\_waiting=1$  there is a good coverage (279), but 
With $num\_waiting=1$ many more states than with other configurations are sent, as every agent sends its $witheld$ states as soon as a single agent is waiting. With $num\_waiting=half$ \mabfws\ obtains a good tradeoff.
In terms of all the considered measures of performance, $num\_waiting=half$ is the best configuration except for the average number of sent messages and the average CPU time. However, for those measures the performance gap to the best configuration is quite limited. 

In Table \ref{tab:resultsfinaliX}, we experimentally evaluate the
 performance of \mabfws\ for different configurations of $who\_send$, that specifies which agents send withheld states.  We
can see that there is no big gap in the performance of these configurations.
For the other experiments in the paper, we use $who\_send=all$ because it performs slightly better in terms of coverage and time, while the average number of exchanged states is pretty
close to the best value obtained for $who\_send=waiting$.

\begin{table}[t]
\centering
\small
\begin{tabular}{|l|c|c|c|c|}
\hline
Performance & $nws$= &$nws$= &$nws$= &$nws$= \\
& $none$ 
& $1$
& $all$
& $group$ \\ \hline 
\bf{Coverage} (320)&258&272&275&{\bf 280} \\ 
Avg time&{\bf 5.05}&5.28&5.22&5.27 \\ 
Avg quality&186.13&186.11&{\bf 185.27}&186.22 \\ 
Avg k-messages&{\bf 79.31}&83.74&85.55&82.72 \\ 
Avg k-states&{\bf 146.67}&180.31&151.72&153.73 \\ 
Time score&241.47&250.44&253.54&{\bf 258.26} \\ 
Quality score&222.2&231.85&236.78&{\bf 238.51} \\ 
Stdev Time&{\bf 0.56}&1.22&1.04&0.85 \\ 
Stdev Quality&15.52&16.02&{\bf 14.9}&15.15 \\ 
	\hline  
\end{tabular}
\caption{Performance of \mabfws\ with $num\_waiting=half$, $who\_send=all$, and $num\_withheld\_states$ (abbreviated with $nws$) set to $1$, $all$, and $group$ in terms of number of solved problems, average CPU time (in seconds), plan cost, number of sent messages (in thousands), number of expanded states (in thousands), time score, quality score, standard deviation of the CPU times, and standard deviations of the plan costs. The best performances are indicated in bold.}\label{tab:resultsfinaliB}
\end{table}

Table \ref{tab:resultsfinaliB} shows the results for different configurations of $num\_withheld\_states$.
With $num\_withheld\_states = none$ the withheld states are not sent; the result is similar to the one obtained with $num\_waiting = all$ (in Table \ref{tab:resultsfinaliS}), confirming the fact that in that case too few states were shared.
By sending one state at a time, i.e., with $num\_withheld\_states = 1$, we have an average number of exchanged messages and an average execution time very close to the configuration for which states are sent in group ($num\_withheld\_states = group$), although eight fewer problems are solved.
Finally, as expected, with $num\_withheld\_states = all$ we have  the highest number of exchanged messages; probably more than necessary given that the performance, in terms of CPU time
and solved problems, is worse than with $num\_withheld\_states = group$. This shows that increasing the number of sent messages has a computational cost, since the average number of expanded states within the time limit used for our experiments is lower than with $num\_withheld\_states$ set to  1 or $group$.

\begin{table}
\centering

\small
\begin{tabular}{|l|c|c|c|}
\hline
Domain & no filtering
& $w^{out}=1$ 
& $w^{out}=2$ \\ \hline 
{\bf From CoDMAP}&&&\\
	\blocksworld&{\bf 20}&{\bf 20}&{\bf 20} \\ 
	\depot&{\bf 20}&18&{\bf 20} \\ 
	\driverlog&{\bf 20}&{\bf 20}&{\bf 20} \\ 
	\elevators&{\bf 20}&{\bf 20}&{\bf 20} \\ 
	\logistics&19&{\bf 20}&{\bf 20} \\ 
	\rovers&{\bf 20}&{\bf 20}&{\bf 20} \\ 
	\satellites&{\bf 20}&{\bf 20}&{\bf 20} \\ 
	\sokoban&14&14&{\bf 17} \\ 
	\taxi&{\bf 20}&{\bf 20}&{\bf 20} \\ 
	\wireless&{\bf 2}&{\bf 2}&{\bf 2} \\ 
	\woodworking&9&{\bf 12}&{\bf 12} \\
	\zenotravel&{\bf 20}&{\bf 20}&{\bf 20} \\ 
	\hline 
{\bf From MBS}&&&\\
	\mabwshort&{\bf 19}&{\bf 19}&{\bf 19} \\ 
	\mabwhardshort&{\bf 15}&{\bf 15}&{\bf 15} \\ 
	\malogshort&19&{\bf 20}&{\bf 20} \\ 
	\maloghardshort&{\bf 20}&{\bf 20}&{\bf 20} \\ 
	\hline 
\hline
Performance & no filtering
& $w^{out}=1$ 
& $w^{out}=2$ \\ \hline 
{\bf From CoDMAP}&&&\\
\bf{Coverage} (320)&277&280&{\bf 285} \\ 
Avg time&16.32&{\bf 6.14}&9.69 \\ 
Avg quality&179.79&{\bf 174.53}&181.26 \\ 
Avg k-messages&1214.2&{\bf 100.6}&547.18 \\ 
Avg k-states&480.18&{\bf 236.85}&323.74 \\ 
Time score &228.73&{\bf 258.69}&245.27 \\ 
Quality score &229.51&{\bf 240.3}&239.25 \\ 
Stdev Time&3.68&{\bf 1.71}&1.97 \\ 
Stdev Quality&17.66&{\bf 14.66}&17.23 \\ 
	\hline 
\end{tabular}
\caption{Coverage (upper table) of \mabfws\ without outgoing novelty filtering, with outgoing novelty $w^{out}$ equal to 1 and 2 for domains of CoDMAP and MBS benchmarks; their performance (bottom table) in terms of number of solved problems, average CPU time (in seconds), plan cost, number of sent messages (in thousands), number of expanded states (in thousands), time score, quality score, standard deviation of the CPU times, and standard deviations of the plan costs. The best performances are indicated in bold.}\label{tab:resultsfinali}
\end{table}

In Table \ref{tab:resultsfinali}, we show the results of \mabfws\ using values of outgoing novelty $w^{out}$ equal to 1 and 2 w.r.t.\ without filtering messages. It is important to remark that the computation and memory cost of determining that the outgoing novelty of a state is $k$ is exponential in $k$, since all the tuples of size up to $k$ but one may be stored and considered. For efficiency, we simplify the computation of the outgoing novelty to only 2 levels for $w^{out} =1$, and only 3 levels for $w^{out}=2$, i.e., the outgoing novelty for  $w^{out}=2$ is determined to be equal to 1, 2, or greater than 2.

The best experimented configuration is $w^{out}=1$. The benefits of this configuration are:
\begin{itemize}
\item A slight improvement of the coverage w.r.t. \mabfws\ without message filtering.
 \item The overall number of exchanged messages is drastically reduced, by $91\%$.
As previously noted, this reduction improves the privacy of planners.
  \item The average execution time is considerably reduced, by almost $50\%$;
  \item There is a substantial increase in the plan quality. This is probably due to the fact that, by holding states with outgoing novelty greater than 1, the priority is given to states that can be reached using a number of actions at most equal to the number of propositions of the problem, i.e., states that in the worst case can be reached by plans shorter than with outgoing novelty greater than 1.
\end{itemize}
Also, \mabfws\ with $w^{out}=2$ leads to a sharp decrease in the average CPU time compared to \mabfws\ without novelty filtering. This is however less than with $w^{out}=1$ for two reasons:
(i) determining that the outgoing novelty for  $w^{out}=1$ is computationally much cheaper than for $w^{out}=2$; (ii) \mabfws\ with $w^{out}=1$ sends fewer states, and probably fewer superfluous states. Compared to \mabfws\ without novelty filtering, the average number of messages exchanged by \mabfws\ with $w^{out}=2$ is lower, and the number of solved problem is higher (eight more than without filtering messages).

\begin{figure}
\scalebox{0.65}{%
    \input{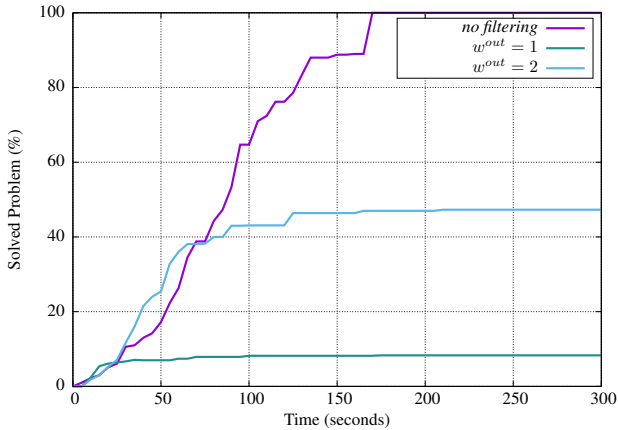}%
    %\unskip
}
\caption{Percentage of  messages sent by \mabfws\ without outgoing novelty filtering, \mabfws\ with outgoing novelty $w^{out}$ equal to 1 and 2 for different CPU-time limits ranging from 0 to 300 seconds. }\label{fig:msgs}
\end{figure}

Figure \ref{fig:msgs} shows the percentage of sent messages w.r.t.\ the total amount of generated messages for the problems solved within different CPU time limits. We can observe that, compared to \mabfws\ without filtering of messages, for CPU time limits greater than 170 seconds, the use of outgoing novelty equal to 1 reduces by one order of magnitude the number of exchanged messages. 
%\medskip

\begin{table}[tb]
\centering

\small
\begin{tabular}{|l|c|c|c|}
\hline
Domain &  {\it secure-}   & $Secure$  & $Secure$ \\
& \mabfws
& $w^{out}=1$ 
& $w^{out}=2$ \\ \hline 
{\bf From CoDMAP}&&&\\
	\blocksworld&{\bf 20}&{\bf 20}&{\bf 20} \\ 
	\depot&{\bf 20}&17&{\bf 20} \\ 
	\driverlog&{\bf 20}&{\bf 20}&{\bf 20} \\ 
	\elevators&{\bf 20}&{\bf 20}&{\bf 20} \\ 
	\logistics&{\bf 20}&{\bf 20}&{\bf 20} \\ 
	\rovers&{\bf 20}&{\bf 20}&{\bf 20} \\ 
	\satellites&{\bf 20}&{\bf 20}&{\bf 20} \\ 
	\sokoban&11&13&{\bf 15} \\ 
	\taxi&{\bf 20}&{\bf 20}&{\bf 20} \\ 
	\wireless&{\bf 2}&{\bf 2}&{\bf 2} \\ 
	\woodworking&11&{\bf 12}&{\bf 12} \\ 
	\zenotravel&{\bf 20}&{\bf 20}&{\bf 20} \\ 
	\hline 
{\bf From MBS}&&&\\
	\mabwshort&{\bf 19}&{\bf 19}&{\bf 19} \\ 
	\mabwhardshort&{\bf 15}&{\bf 15}&{\bf 15} \\ 
	\malogshort&{\bf 20}&{\bf 20}&{\bf 20} \\ 
	\maloghardshort&19&{\bf 20}&{\bf 20} \\ 
	\hline 
	\hline 
	 & {\it secure-}  & $Secure$  & $Secure$ \\
Performance & \mabfws
& $w^{out}=1$ 
& $w^{out}=2$ \\ \hline 
\bf{Coverage} (320)&277&278&{\bf 283} \\ 
Avg time&15.55&{\bf 5.79}&10.42 \\ 
Avg quality&182.3&{\bf 172.56}&181.85 \\ 
Avg k-messages&1036.47&{\bf 87.15}&525.73 \\ 
Avg k-states&412.9&{\bf 184.23}&304.79 \\ 
Time score &230.52&{\bf 257.88}&240.34 \\ 
Quality score&233.76&{\bf 240.76}&239.73 \\ 
Stdev Time&3.36&{\bf 1.37}&1.79 \\ 
Stdev Quality&16.27&16.3&{\bf 16.06} \\ 
	\hline 

\end{tabular}
\caption{\label{tab:secure}  
Coverage (upper table) of \mabfwssecure\ without outgoing novelty filtering, with outgoing novelty $w^{out}$ equal to 1 and 2 for domains of CoDMAP and MBS benchmarks; their performance (bottom table) in terms of number of solved problems, average CPU time (in seconds), plan cost, number of sent messages (in thousands), number of expanded states (in thousands), time score, quality score, standard deviation of the CPU times, and standard deviations of the plan costs. The best performances are indicated in bold.}
\end{table}

Table \ref{tab:secure} shows the performance of \mabfws\ with the ``secure check'' proposed by \citeauthor{brafman15privacy} (\citeyear{brafman15privacy}) for the exchanged messages, denoted by \mabfwssecure. This version is incomplete but strong privacy-preserving over domains such as \logistics, \rovers\ and \satellites, where the heuristic function is independent from the private part of the problem \cite{brafman15privacy}, as it does not send states whose public projection has been sent before. Comparing the results with Table \ref{tab:resultsfinali}, we can
 observe a very limited decrease in terms of coverage and an improvement
 in terms of transmitted messages and expanded states.

\begin{table}
\centering
\scriptsize
%\hspace{-5cm}
\begin{tabular}{|l|c|c|c|c|}
\hline
Domain & \mabfws & \masiw &  \maplan & \psm \\
&  $w^{out} = 1$ & & & \\
\hline
	\blocksworld &{\bf 20} &{\bf 20} &{\bf 20} &{\bf 20} \\ 
	\depot       &{17}        & 8     &12       & 17\\ 
	\driverlog   &{\bf 20} &{\bf 20} &16       &{\bf 20} \\ 
	\elevators   &{\bf 20} &{\bf 20} &8        & 12\\ 
	\logistics   &{\bf 20}        & 18  &18       & 18\\ 
	\rovers      &{\bf 20} &{\bf 20} &{\bf 20} & 19 \\ 
	\satellites  &{\bf 20} &{\bf 20} &{\bf 20} & 13\\ 
	\sokoban     &{ 13}        & 4   &{\bf 17} & 16 \\ 
	\taxi        &{\bf 20} &{\bf 20} &{\bf 20} &{\bf 20}\\ 
	\wireless    &2        &0        &{\bf 4}  & 0 \\ 
	\woodworking &12        &1       &{ 15}    &{\bf 18} \\ 
	\zenotravel  &{\bf 20} &{\bf 20} &{\bf 20} & 10 \\ 
	\hline 
{\bf Overall} (240)&204   &171      &190      & 184 \\ 
	\hline 
\end{tabular}%\vspace*{-0.1cm}
\caption{\label{tab:StateOfTheArt}Number of problems solved by  \mabfws\ with outgoing novelty filtering  $w^{out}$ equal to 1 w.r.t.\ \masiw, \maplan\ and \psm\ for the benchmark problems of CoDMAP.  The best performances are indicated in bold.}% \vspace*{-.4cm}}
\end{table}

We also compare our approach with other three existing approaches, \masiw, the approach mostly related to our work, \psm, and the best performing configuration of \maplan. The latter two planners were the best  planners that took part in the CoDMAP competition. Table \ref{tab:StateOfTheArt} shows the results of this comparison for the CoDMAP domains. As for benchmark MBS, \masiw\ solves no problem, while \maplan\ and \psm\ do not support private goals, which are present in these problems. The time limit used for this comparison is 30 minutes, the same limit as in the competition. The results in Table \ref{tab:StateOfTheArt} show that for the competition problems \mabfwssecure\ with $w^{out}=1$ outperforms \masiw\ and is better than \maplan\ and \psm. Remarkably, the only type of information that agents share by using our approach is related to the exchanged search states, while \maplan\ also requires sharing the information for the computation of the search heuristics. In this sense, besides solving a larger set of problems, \mabfws\ exposes less private knowledge to other agents.%\vspace*{-0.2cm}

%\vspace*{3mm}
\subsection{Messages delay}
In this section, we present an experiment aimed at testing how \mabfws\ with filtering of messages performs in a heavily
congested distributed network. With this aim, we introduced a mechanism that, by means of an artificial delay  applied on each exchanged message, can simulate arbitrarily network delays during  message transmission.
These delays are distributed according to the gamma distribution that is an approximation of the delays in the Internet network \cite{DBLP:journals/corr/abs-0907-4468}. 
In particular, Table \ref{tab:resultsDelayBig} considers 5 different configurations of delays, with a standard deviation equal to 10\% of the average delay.

\begin{table}[tb]
\scriptsize
\begin{center}
\begin{tabular}{|l|c|c|c|c|c|}
\hline
Domain & no delay
& 10ms
& 100ms
& 1s
& 10s \\ \hline 
{\bf From CoDMAP}&&&&&\\
	\blocksworld&{\bf 20}&{\bf 20}&{\bf 20}&{\bf 20}&19 \\ 
	\depot&18&{\bf 19}&{\bf 19}&9&5 \\ 
	\driverlog&{\bf 20}&{\bf 20}&{\bf 20}&{\bf 20}&19 \\ 
	\elevators&{\bf 20}&{\bf 20}&{\bf 20}&{\bf 20}&19 \\ 
	\logistics&{\bf 20}&{\bf 20}&{\bf 20}&18&15 \\ 
	\rovers&{\bf 20}&{\bf 20}&{\bf 20}&{\bf 20}&{\bf 20} \\ 
	\satellites&{\bf 20}&{\bf 20}&{\bf 20}&{\bf 20}&{\bf 20} \\ 
	\sokoban&{\bf 14}&13&{\bf 14}&12&12 \\ 
	\taxi&{\bf 20}&{\bf 20}&18&0&0 \\ 
	\wireless&{\bf 2}&{\bf 2}&{\bf 2}&1&0 \\ 
	\woodworking&11&12&{\bf 13}&{\bf 13}&11 \\ 
	\zenotravel&{\bf 20}&{\bf 20}&{\bf 20}&19&{\bf 20} \\ 
	\hline 
{\bf From MBS}&&&&&\\
	\mabwshort&{\bf 19}&{\bf 19}&{\bf 19}&{\bf 19}&{\bf 19} \\ 
	\mabwhardshort&{\bf 15}&{\bf 15}&{\bf 15}&{\bf 15}&{\bf 15} \\ 
	\malogshort&{\bf 20}&{\bf 20}&{\bf 20}&{\bf 20}&0 \\ 
	\maloghardshort&{\bf 20}&{\bf 20}&{\bf 20}&{\bf 20}&0 \\ 
	\hline 
	\hline
	Performance & no delay
& 10ms
& 100ms
& 1s
& 10s \\ \hline 
\bf{Coverage} (320)&279&{\bf 280}&{\bf 280}&246&194 \\ 
Avg time&{\bf 4.89}&5.0&6.56&22.14&162.92 \\ 
Avg quality&153.13&148.28&141.27&138.52&{\bf 136.26} \\ 
Avg k-messages&{\bf 17.94}&23.45&19.11&57.82&651.98 \\ 
Avg k-states&{\bf 62.14}&71.59&92.76&365.39&1374.01 \\ 
Time score&{\bf 265.14}&261.1&215.12&127.06&68.48 \\ 
Quality score&234.17&241.08&{\bf 252.83}&218.65&167.95 \\ 
Stdev Time&0.88&0.65&{\bf 0.62}&0.92&5.89 \\ 
Stdev Quality&12.7&11.08&8.17&{\bf 7.18}&7.98 \\ 
	\hline 

\end{tabular}
\end{center}
\caption{
Coverage (upper table) of \mabfws\ with no delay in the transmission of messages, with an average delay in the transmission of each message equal to 10 ms, 100ms, 1s, and 10s for domains of CoDMAP and MBS benchmarks; their performance (bottom table) in terms of number of solved problems, average CPU time (in seconds), plan cost, number of sent messages (in thousands), number of expanded states (in thousands), time score, quality score, standard deviation of the CPU times, and standard deviations of the plan costs. The best performances are indicated in bold.}\label{tab:resultsDelayBig}
\end{table}

For average delays smaller than $100$ ms, in terms of coverage,  we observe no significant  performance gap w.r.t. \mabfws\ with no delay in the transmission of messages.
Instead, as one could expect, with very high delays the number of solved problems decreases. However, even with an average delay of 10 seconds applied to each exchanged message, the number of solved problems does not decrease to few units. This probably indicates that many problems in our benchmarks does not require an intensive cooperation among agents. 

Furthermore, we can observe a significant increase of the execution time when the delay increases. The required CPU time varies from $4.89$ seconds for \mabfws\ without delay, to $162.92$ seconds for the 
highest average delay we considered in our experiment. The higher number of expended states  when the average delay is high is probably due to the fact that, in that case, the agents can spend more time for the state expansion.
Moreover, concerning the plan quality, contrary to what one could expect,
the best value is found with the greatest delay.
A possible explanation of this behavior is related to the high number of expanded states, which helps
to find better plans.

%\vspace*{3mm}
\section{Conclusion}

In this paper, we show that novelty based techniques are very useful to significantly reduce the number of messages transmitted among agents, better preserving their privacy levels as well as improving their performance.
The experimental analysis in this paper shows the effectiveness of our techniques in terms of coverage, CPU-time and number of transmitted messages. Moreover, we observed that our approach is robust to delays in the transmission of messages that could occur in overloaded networks.

In future work we intend to explore the usage of privacy-preserving set operations for secure multi-party computation \cite{kissner2005privacy}, and the secure computation of joint novelty functions among agents. This may lead \mabfws\ to coordinate agents through a joint novelty table.

\section*{Acknowledgements}
This research carried out with the support of resources of the National Collaborative Research Infrastructure Strategy (NeCTAR), and the Big \& Open Data Innovation Laboratory (BODaI-Lab) of the University of Brescia, which is granted by Fondazione Cariplo and Regione Lombardia. Nir Lipovetzky, has been partially funded by DST group.

\bibliographystyle{aaai}
\bibliography{biblio}

\end{document}